\documentclass{article}

\PassOptionsToPackage{round}{natbib}

\usepackage[final]{neurips_2024}

\usepackage[utf8]{inputenc} 
\usepackage[T1]{fontenc}    
\usepackage{hyperref}       
\usepackage{url}            
\usepackage{booktabs}       
\usepackage{amsfonts}       
\usepackage{nicefrac}       
\usepackage{microtype}      
\usepackage{xcolor}         

\usepackage{amsmath,amsthm,amssymb}
\usepackage{doi}
\usepackage{cleveref}
\usepackage{tikz}
\usetikzlibrary{shapes.callouts} 
\usetikzlibrary{arrows,decorations.pathmorphing,backgrounds,calc}
\usetikzlibrary{shapes.callouts} 
\usetikzlibrary{shapes.geometric}
\usetikzlibrary{positioning,automata,arrows}
\usepackage{tikzsymbols}
\usepackage{wrapfig}
\usepackage{xspace}

\newcommand{\emoji}[1]{}

\definecolor{color1}{RGB}{230,97,0}
\definecolor{color2}{RGB}{93,58,155}
\definecolor{color3}{HTML}{90A4AE}
\definecolor{color4}{HTML}{EEEEEE}

\tikzset{
  speech/.style  = { draw, rectangle callout, callout relative pointer={#1}},
  thought/.style  = { draw, cloud callout, callout relative pointer={#1}, cloud puffs = 30, cloud puff arc = 120}
}

\usepackage{caption}
\usepackage{subcaption}

\newcommand{\tempagg}{W_\text{ex}}
\newcommand{\reals}{\mathbb{R}}

\theoremstyle{definition}
\newtheorem{definition}{Definition}

\newcommand{\tuple}[1]{\langle #1 \rangle}

\newcommand{\schemename}{temporal pluralism\xspace}
\newcommand{\Schemename}{Temporal pluralism\xspace}
\newcommand{\scorename}{\schemename}
\newcommand{\Scorename}{\Schemename}

\title{Pluralistic Alignment Over Time}

\author{%
  Toryn Q. Klassen, Parand A. Alamdari, Sheila A. McIlraith \\
  University of Toronto \& Vector Institute \\
  Schwartz Reisman Institute for Technology and Society\\
  Toronto, Ontario, Canada\\
  \texttt{\{toryn,parand,sheila\}{@cs.toronto.edu}} 
}

\begin{document}

\begin{figure}[b]
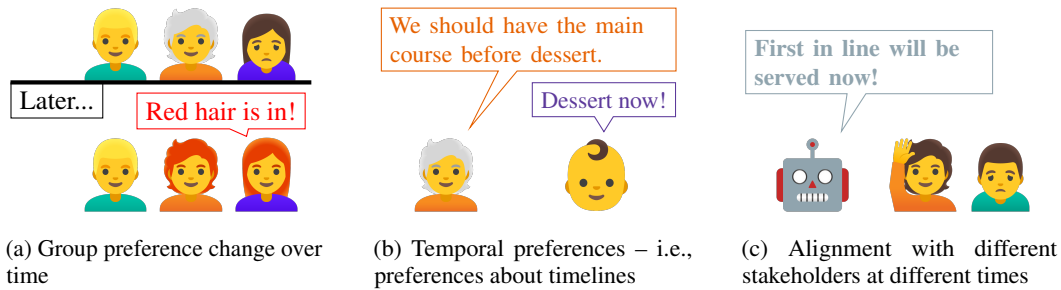

    \centering
    \vspace{-0.1em}
    \begin{subfigure}[b]{0.3\textwidth}
         \centering
         \input{figs/fig-change}
         \caption{Group preference change over time}
         \label{fig:c}
     \end{subfigure}
    \hfill
    \begin{subfigure}[b]{0.3\textwidth}
         \centering
         \input{figs/fig-temporal-prefs}
         \caption{Temporal preferences -- i.e., preferences about timelines}
         \label{fig:b}
     \end{subfigure}
     \hfill
     \begin{subfigure}[b]{0.3\textwidth}
         \centering
         \input{figs/fig-times}
         \caption{Alignment with different stakeholders at different times}
         \label{fig:a}
     \end{subfigure}
    \caption{Different temporal aspects of pluralistic alignment}
    \label{fig:aspects}
\end{figure}
\maketitle

\begin{abstract}
If an AI system makes decisions over time, how should we evaluate how aligned it is with a group of stakeholders (who may have conflicting values and preferences)? In this position paper, we advocate for consideration of temporal aspects including stakeholders' changing levels of satisfaction and their possibly temporally extended preferences. We suggest how a recent approach to evaluating fairness over time could be applied to a new form of pluralistic alignment: temporal pluralism, where the AI system reflects different stakeholders' values at different times.
\end{abstract}

\section{Temporal Aspects of Pluralistic Alignment}

In this paper we consider aligning an AI system's decisions with the values or preferences of multiple stakeholders, a topic that has been recently drawing attention \cite[e.g.,][]{SorensenICML2024roadmap,ConitzerICML2024social}, in the context of sequential decision making.
\citeauthor{SorensenICML2024roadmap} wrote that
\begin{quote}
     [W]e need
systems that are \emph{pluralistic}, or capable of representing a
diverse set of human values and perspectives.
\end{quote}

Motivated by our context of \emph{sequential} decision making, which takes place over time, 
we focus on \emph{temporal} aspects of alignment.
There are a variety of ways that time may factor into evaluating how aligned an AI system is, including (as illustrated in \Cref{fig:aspects})
\begin{enumerate}
    \item preference change over time,
    \item temporally extended preferences,
    \item and how pluralistic alignment may only be realizable over time, through acting to achieve different stakeholder's interests at different times.
\end{enumerate}
Each of these points will be elaborated upon in the following subsections (and we encourage further research on them), but
the third will then be the focus of the rest of this paper. We will adapt our recently introduced framework for temporally extended \emph{fairness} \citep{AlamdariICML2024remembering} to pluralistic alignment over time.

\subsection{Preference change over time}

The preferences of a society or group may change over time. One of the ways of operationalizing pluralistic alignment suggested by \citet{SorensenICML2024roadmap} (for LLMs) was \emph{Overton pluralism}, where the LLM responds to a query by giving all answers in the Overton window. However, the Overton window is often spoken about as shifting in reality \citep[e.g.,][]{AstorNYT2019mainstream}. Furthermore, actions taken by AI systems could potentially influence such shifts, e.g., by changing enough individual preferences \citep{CarrollICML2024changing} or by influencing demographic changes. Less diverse preferences might be easier to satisfy, which could give an AI system an incentive for manipulation.

 \subsection{Temporally extended preferences, norms, goals, and rewards}
\label{sec:nonmarkov}

People may also have explicitly temporal preferences, i.e., preferences about how the state evolves over time. (There is a body of work in AI planning about satisfying (typically one agent's) temporally extended goals and preferences \citep[e.g.,][]{BacchusAMAI1998extended,SonLPNMR2004preferences,BaierAIJ2009preferences,BienvenuAIJ2011preferred}.)

To illustrate, one might prefer that dessert be served after the main course, while someone else might want to eat dessert first (\Cref{fig:b}). What order should the meal's courses be served in? For this case, there may also be social norms or conventions to consider. \citet{ZhiXuan2024beyond} have suggested that AI systems should be aligned with norms rather than aggregated preferences. Norms may also be temporally extended \citep[e.g.,][]{PorfirioUIST2018authoring,KasenbergAIES2018norms,MalleROMAN2023norms}.

{
\newcommand{\dessert}{\raisebox{-0.2em}{\includegraphics[width=1em]{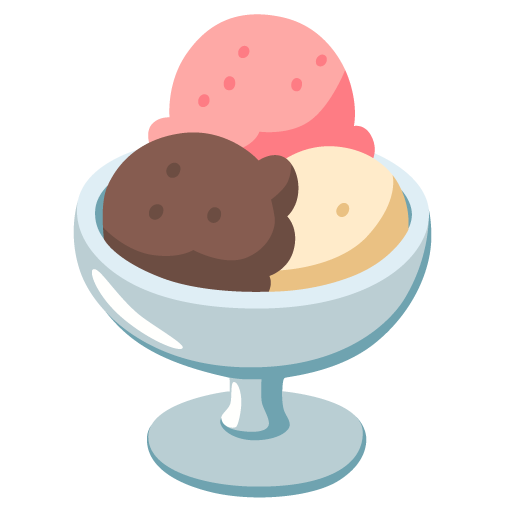}}}
\newcommand{\maincourse}{\raisebox{-0.2em}{\includegraphics[width=1em]{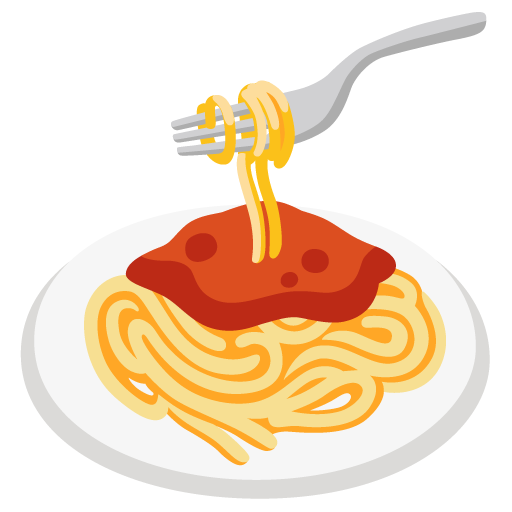}}}
\begin{wrapfigure}[19]{r}{0.44\textwidth}
    \centering
    \begin{tikzpicture}[node distance=2cm,on grid,every initial by arrow/.style={ultra thick,->, >=stealth}]
  \node[ultra thick,state,initial left,fill=color4] (q_0) at (0,0) {$u_0$};
  \node[ultra thick,state,fill=color4]         (q_1) at (3.5, 0)  {$u_1$};
   \node[ultra thick,state,fill=color4]         (q_2) at (1.75, -2)  {$u_2$};

  \path[ultra thick,->, >=stealth] (q_0) edge node [above] {$\tuple{\maincourse\wedge\neg\dessert,0}$} (q_1);

  \path[ultra thick,->, >=stealth] (q_1) edge node [right] {$\ \tuple{\dessert,1}$} (q_2);

  \path[ultra thick,->, >=stealth] (q_0) edge [loop above] node {$\tuple{\neg\maincourse\wedge \neg\dessert ,0}$} ();

  \path[ultra thick,->, >=stealth] (q_0) edge node [left] {$\tuple{\dessert,0}\ $} (q_2);

  \path[ultra thick,->, >=stealth] (q_1) edge [loop above] node {$\tuple{\neg\dessert ,0}$} ();


  \path[ultra thick,->, >=stealth] (q_2) edge [loop above] node {$\tuple{\top ,0}$} ();

\end{tikzpicture}
    \caption{A reward machine that gives reward 1 only on trajectories on which dessert (\dessert) is eaten after the main course (\maincourse). An edge labelled $\tuple{\varphi, r}$ is taken when the propositional formula $\varphi$ is true, yielding reward $r$.}
    \label{fig:rm}
\end{wrapfigure}
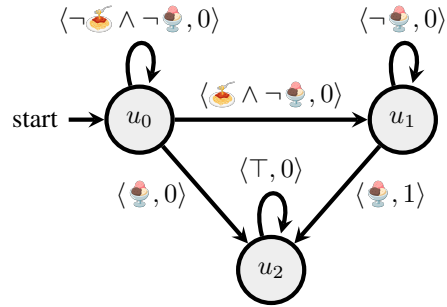

In automated sequential decision making, it's common (for example, when using Markov Decision Processes) to represent preferences or other things to be optimized using a reward function 
that assigns a numeric value $R(s_t,a_t,s_{t+1})$ to each transition from a state $s_t$ to another state $s_{t+1}$ using action $a_t$.
However, it's also possible to define a \emph{non-Markovian} reward function $R(\tau)$ that maps a whole trajectory of alternating states and actions $\tau=s_1,a_1,s_2,\dots, a_T,s_{T+1}$ to  a real number. This allows for rewarding temporally extended behaviors.

Non-Markovian reward functions have been describing using temporal logics including LTL \cite[e.g.,][]{BacchusAAAI1996behaviors,ThiebauxJAIR2006nonmarkovian,CamachoSOCS2017shaping} and reward machines \citep{ToroIcarteICML2018machines,ToroIcarteJAIR2022machines,CamachoIJCAI2019beyond}. An example reward machine is shown in \Cref{fig:rm}. 
In the context of alignment, \citet{ZhiXuan2024beyond} recently encouraged the consideration of temporal logics and reward machines to better represent human preferences.

Another temporal aspect of preferences is temporal discounting, where future rewards are valued less than current ones. \cite{PitisNeurIPS2023consistent} 
considers aggregating the preferences of multiple stakeholders who each have a Markovian reward function, but possibly different discount factors, and argues that in general the aggregation should be a non-Markovian reward function.
Finally, we note that the \emph{dynamic reward MDPs} used by
\citet{CarrollICML2024changing} to model changing preferences also resemble reward machines.

}

\subsection{Pluralistic alignment over time}

{Pluralistic alignment requires consideration of a collection of human preferences, but such preferences may  be conflicting, such that after any one decision, alignment is only with a subset of humans. In a sequential decision making setting, such disparities can be mitigated by future actions. More generally,} the decisions made by AI (or human) systems will often cause different people to be better off at different times. How should such temporal tradeoffs be evaluated to determine how good the decisions were in the short, medium and long term \citep{AlamdariICML2024remembering}? To illustrate, consider the problem of distributing goods among people -- logistic constraints will typically prevent everyone from getting the same amount of goods at exactly the same time.  
On a longer timescale, how the utilities of different generations of people (living at different times) should be aggregated is a question that has been considered in economics \citep[e.g.,][]{ZameTE2007intergenerational,AlvarezJEEM2009mixed}.
    
Indeed, at different times, an AI system may seem to be better aligned with different stakeholders. More positively, while it may not be possible to satisfy everyone with a single decision, a sequence of decisions can reflect a diversity of values. This suggests a new way that a system could be pluralistic: \textbf{temporal pluralism}, wherein the system reflects different stakeholders' values at different times.

Which stakeholders' values should be reflected at which times?

The simplest answer might be just that each stakeholder should be satisfied a fair \emph{fraction} of the time.
That is closely related to the notion of \emph{distributional pluralism} from \cite{SorensenICML2024roadmap}.
Focusing on alignment for language models,
\citeauthor{SorensenICML2024roadmap} defined a distributionally pluralistic language model $\mathcal{M}$ as having the property that
\begin{quote}
    For a given prompt $x$, $\mathcal{M}$ is as
likely to provide response $y$ as the reference population $G$.
\end{quote}
Assume the population provides responses it prefers, and a subgroup of $n$\% of the population prefers a particular sort of response (and the rest of the population prefers other responses). Then if a distributionally pluralistic model is queried repeatedly, in expectation $n$\% of its responses would be preferred by the subgroup.

However, some time periods may be viewed as more important than others (because of temporal discounting or other reasons), which suggests a potential need for a more structured approach (e.g., something like turn-taking). Furthermore, in the context of sequential decision making, constraints may be imposed by what state transitions are physically possible, and the AI system may need to plan ahead to be able to satisfy different stakeholders in the future. The matter may be further complicated by stakeholders having temporally extended preferences about the overall timeline, as we discussed in \Cref{sec:nonmarkov}.

\section{A framework for pluralistic alignment over time}

In this section we adapt the framework for temporally extended fairness from our previous work \citep{AlamdariICML2024remembering}\footnote{{Another recent approach to fairness over time was introduced by \cite{TorresECAI2024temporal}.}} to apply to the problem of pluralistic alignment.

\paragraph{Notation} When writing a trajectory of states and actions $\tau_{T}=s_1,a_1,s_2,\dots, a_T,s_{T+1}$, the subscript $T$ on $\tau$ (if present) indicates the number of actions in the trajectory. Given $\tau_T$, we can write $\tau_i$ (where $i<T$) for the prefix of $\tau_{T}$ ending with $s_{i+1}$.

What we want is to be able to evaluate how well a trajectory of states and actions reflects the preferences of a collection of stakeholders.
We previously had the notion of a \emph{fairness scheme} \cite[Def. 4.5]{AlamdariICML2024remembering}, which we adopt as a \emph{\schemename scheme}:
\begin{definition}[\Schemename scheme]
    Given a state space $S$ and action space $A$, a \schemename scheme for $n$ agents is a tuple 
    $\tuple{U,\tempagg,B}$ where 
    \begin{itemize}
        \item $U: (S\times A)^*\times S \to \reals^n$ is the stakeholder status function.
        \item $\tempagg:(\reals^n)^*\to\reals$ is the extended aggregation function.
        \item $B:(S\times A)^*\times S\to\{0,1\}$ is the filter function.
    \end{itemize}
\end{definition}

$U$'s output, a vector of length $n$, is meant to measure how ``good'' the input (a trajectory $\tau$ of states and actions) has been for each of $n$ stakeholders. 
For example, the measure could be, for each stakeholder, the (discounted) sum of that stakeholder's rewards over the trajectory. \citeauthor{AlamdariICML2024remembering} assumed the $i$th stakeholder has a (Markovian) reward function $R_i(s,a,s')$, but we note that a \emph{non-Markovian} reward function $R_i(\tau)$ is just as compatible with the approach. 

{We use the extended aggregation function to compute a \emph{temporal pluralism score}} by aggregating the outputs of the stakeholder status function over time. The filter function is used to restrict which time points are considered by the extended aggregation function.
We adopt 
\cite[Def. 4.7]{AlamdariICML2024remembering} as a \emph{\scorename score}:
\begin{definition}[\Scorename score]
Given a trajectory $\tau_T=s_1, a_1, s_2, \dots, a_T, s_{T+1}$, the \scorename score of $\tau_T$ according to the \schemename scheme $\tuple{U,\tempagg,B}$ is
$$\tempagg(U(\tau_{t_1}),U(\tau_{t_2}),\dots, U(\tau_{t_k}))$$
where $(t_1,t_2,\dots,t_k)$ is the subsequence of $(1,2,\dots,T)$ for which $B(\tau_{t_i})=1$ for each $i$.%
\end{definition}

{The extended aggregation function can be thought of as a temporally extended social welfare}, which looks at how well each stakeholder is doing at each given point in time.

For a simple illustration, consider a scenario (inspired by \cite{LacknerAAAI2020perpetual}) where an AI assistant books restaurants for the 
{frequent} joint dinners of a
{group of five friends}
with diverse culinary preferences.
We could evaluate the AI using a \schemename scheme with the following components:
\begin{itemize}
    \item $U(\tau)$ includes, for each stakeholder (member of the friend group), how often they went to a restaurant of their preferred type over the course of the trajectory $\tau$.
    \item {$\tempagg(u_1,u_2,\dots,u_k)=\mathit{Nash}(u_{11},\dots, u_{1n},u_{21},\dots,u_{2n},\dots, u_{k1},\dots, u_{kn})$ where $\mathit{Nash}$ is Nash welfare, a standard social welfare function (whose value is just the product of its inputs) and $u_{ij}$ is the $j$th entry of the vector $u_i$. That is, we compute the Nash welfare as though each temporal version of each stakeholder were another individual.}\footnote{Other options would include having $\tempagg$ first aggregate across temporal versions of each stakeholder and then aggregate across stakeholders, or vice versa.}
    \item $B(\tau)=1$ only on those time steps on which another
    {ten}
    restaurants have been visited.
\end{itemize}
{The idea is to give higher scores to trajectories on which not only have a variety of restaurants been visited in the long term, but also during the process (for each
{ten} restaurants).}

As \citeauthor{AlamdariICML2024remembering} noted, 
\schemename schemes allow for \emph{long-term}, \emph{periodic}, and \emph{anytime} evaluations.
\begin{description}
    \item[Long-term] The score given by a long-term \schemename scheme to a trajectory $\tau_T$ only varies with the last stakeholder status $U(\tau_T)$. This ignores the evolution before the end (except insofar as the stakeholder status captures it).
    \item[Periodic] A periodic \schemename scheme (with period $p$) ignores or filters out times that are not a multiple of $p$, so that the \scorename score is equal to $\tempagg(U(\tau_p),U(\tau_{2p}),\dots, U(\tau_{\lfloor T/p \rfloor p}))$. This might be desirable, for example, in a situation where a robot is distributing goods to a group of people, and it takes a while for each round of deliveries to be completed (before the end of a round, it might seem like some people are being ignored, and so that the robot is not aligned with them).
    \item[Anytime] An anytime \schemename scheme is a periodic scheme with period 1. Depending on the extended aggregation function $\tempagg$, achieving a high \scorename score with such a scheme may be difficult or impossible.
\end{description}

An AI system trying to act so as to bring about a trajectory with a high \scorename score would in general need to be able to \emph{remember} some information about the past.
\citeauthor{AlamdariICML2024remembering} presented a reinforcement learning approach using an explicit memory, which was able to deal with certain types of fairness schemes, but further work is needed.

\section{Conclusion}

We have argued for further consideration of temporal aspects of alignment with multiple stakeholders, and suggested that in some cases it may only be possible to achieve pluralistic alignment, reflecting a diversity of preferences or values, \emph{over time}.
We further suggested adapting the approach to temporally extended fairness from \cite{AlamdariICML2024remembering} to the problem of pluralistic alignment.
Further work is needed to investigate what specific \schemename schemes would be most appropriate for aggregating people's preferences. Frameworks that people use to organize the satisfaction of their interests over time, like turn-taking and queuing, may be informative in making this choice.

\section*{Acknowledgements}

We wish to acknowledge funding from the Natural Sciences and Engineering Research Council of Canada (NSERC) and the Canada CIFAR AI Chairs Program (Vector Institute). 
The first author also received funding from Open Philanthropy.
{Resources used in preparing this research were provided, in part, by the Province of
Ontario, the Government of Canada through CIFAR, and
companies sponsoring the Vector Institute for Artificial Intelligence.}

\bibliographystyle{plainnat}
\bibliography{ref}

\end{document}